\documentclass[sigconf]{acmart}

\definecolor{stype1}{HTML}{14BBE2}
\definecolor{scvd}{HTML}{9F2332}
\definecolor{sred}{HTML}{B02418}

\AtBeginDocument{%
  \providecommand\BibTeX{{%
    \normalfont B\kern-0.5em{\scshape i\kern-0.25em b}\kern-0.8em\TeX}}}

\copyrightyear{2024}
\acmYear{2024}
\setcopyright{acmlicensed}
\acmConference[KDD '24] {Proceedings of the 30th ACM SIGKDD Conference on Knowledge Discovery and Data Mining }{August 25--29, 2024}{Barcelona, Spain.}
\acmBooktitle{Proceedings of the 30th ACM SIGKDD Conference on Knowledge Discovery and Data Mining (KDD '24), August 25--29, 2024, Barcelona, Spain}
\acmISBN{979-8-4007-0490-1/24/08}
\acmDOI{10.1145/3637528.3671594}

\usepackage{xspace}
\newcommand{\methodname}{TACCO\xspace}
\usepackage{multirow}
\usepackage{bm}
\usepackage{graphicx}
\usepackage{subcaption}
\usepackage{pifont}
\usepackage{xcolor}
\usepackage{amsmath}
\usepackage{balance}
\usepackage{colortbl} 
\usepackage{enumitem}
\usepackage{mathtools}
\usepackage{titlesec}
\usepackage{caption}

\titlespacing*{\section}{0pt}{5pt}{2pt}
\titlespacing*{\subsection}{3pt}{3pt}{0pt}

\titleformat{\section}{\Large\bfseries\MakeUppercase}{\thesection}{1em}{}

\begin{document}

\title[TACCO: Task-guided Co-clustering of Clinical Concepts and Patient Visits for Disease Subtyping on EHR]{TACCO: Task-guided Co-clustering of Clinical Concepts and Patient Visits for Disease Subtyping based on EHR Data}


\author{Ziyang Zhang}
\affiliation{%
  \institution{Emory University}
  \streetaddress{400 Dowman Drive}
  \city{Atlanta}
  \state{Georgia}
  \country{USA}
  \postcode{30322}}
\email{ziyang.zhang2@emory.edu}

\author{Hejie Cui}
\affiliation{%
  \institution{Emory University}
  \streetaddress{400 Dowman Drive}
  \city{Atlanta}
  \state{Georgia}
  \country{USA}
  \postcode{30322}}
\email{hejie.cui@emory.edu}

\author{Ran Xu}
\affiliation{%
  \institution{Emory University}
  \streetaddress{400 Dowman Drive}
  \city{Atlanta}
  \state{Georgia}
  \country{USA}
  \postcode{30322}}
\email{ran.xu@emory.edu}

\author{Yuzhang Xie}
\affiliation{%
  \institution{Emory University}
  \streetaddress{400 Dowman Drive}
  \city{Atlanta}
  \state{Georgia}
  \country{USA}
  \postcode{30322}}
\email{yuzhang.xie@emory.edu}

\author{Joyce C. Ho}
\affiliation{%
  \institution{Emory University}
  \streetaddress{400 Dowman Drive}
  \city{Atlanta}
  \state{Georgia}
  \country{USA}
  \postcode{30322}}
\email{joyce.c.ho@emory.edu}

\author{Carl Yang}
\authornote{Carl Yang is the corresponding author.}
\affiliation{%
  \institution{Emory University}
  \streetaddress{400 Dowman Drive}
  \city{Atlanta}
  \state{Georgia}
  \country{USA}
  \postcode{30322}}
\email{j.carlyang@emory.edu}

\renewcommand{\shortauthors}{Zhang, et al.}

\begin{abstract}
The growing availability of well-organized Electronic Health Records (EHR) data has enabled the development of various machine learning models towards disease risk prediction. However, existing risk prediction methods overlook the heterogeneity of complex diseases, failing to model the potential disease subtypes regarding their corresponding patient visits and clinical concept subgroups. In this work, we introduce \textbf{\methodname}, a novel framework that jointly discovers clusters of clinical concepts and patient visits based on a hypergraph modeling of EHR data. Specifically, we develop a novel self-supervised co-clustering framework that can be guided by the risk prediction task of specific diseases. Furthermore, we enhance the hypergraph model of EHR data with textual embeddings and enforce the alignment between the clusters of clinical concepts and patient visits through a contrastive objective. Comprehensive experiments conducted on the public MIMIC-III dataset and Emory 
internal CRADLE dataset over the downstream clinical tasks of phenotype classification and cardiovascular risk prediction demonstrate an average 31.25\% performance improvement compared to traditional ML baselines and a 5.26\% improvement on top of the vanilla hypergraph model without our co-clustering mechanism. In-depth model analysis, clustering results analysis, and clinical case studies further validate the improved utilities and insightful interpretations delivered by \textbf{\methodname}. Code is available at \textcolor{blue}{\url{https://github.com/PericlesHat/TACCO}}.

\end{abstract}



\maketitle

\section{INTRODUCTION}
Electronic Health Records (EHR) is a significant advancement in medical data management. They store patient data such as medical history, treatments, and lab results. EHRs have played a crucial role in advancing healthcare applications, such as treatment decision support \cite{sutton2020overview} and preventive care \cite{liu2018deep, shah2017extracting}.
In recent years, machine learning (ML) has been employed to extract valuable insights from EHRs and is widely studied in healthcare informatics. This leads to innovations in various applications such as suicide risk prediction \cite{cheng2016risk, su2020machine}, diagnosis prediction \cite{pang2021prediction, ma2017dipole, hosseini2018heteromed}, phenotypes classification \cite{fu2019ddl, xu2022counterfactual, xu2023hypergraph}, and drug recommendation \cite{yang2021safedrug}.

\begin{figure}[t]
  \centering
  \includegraphics[width=\linewidth]{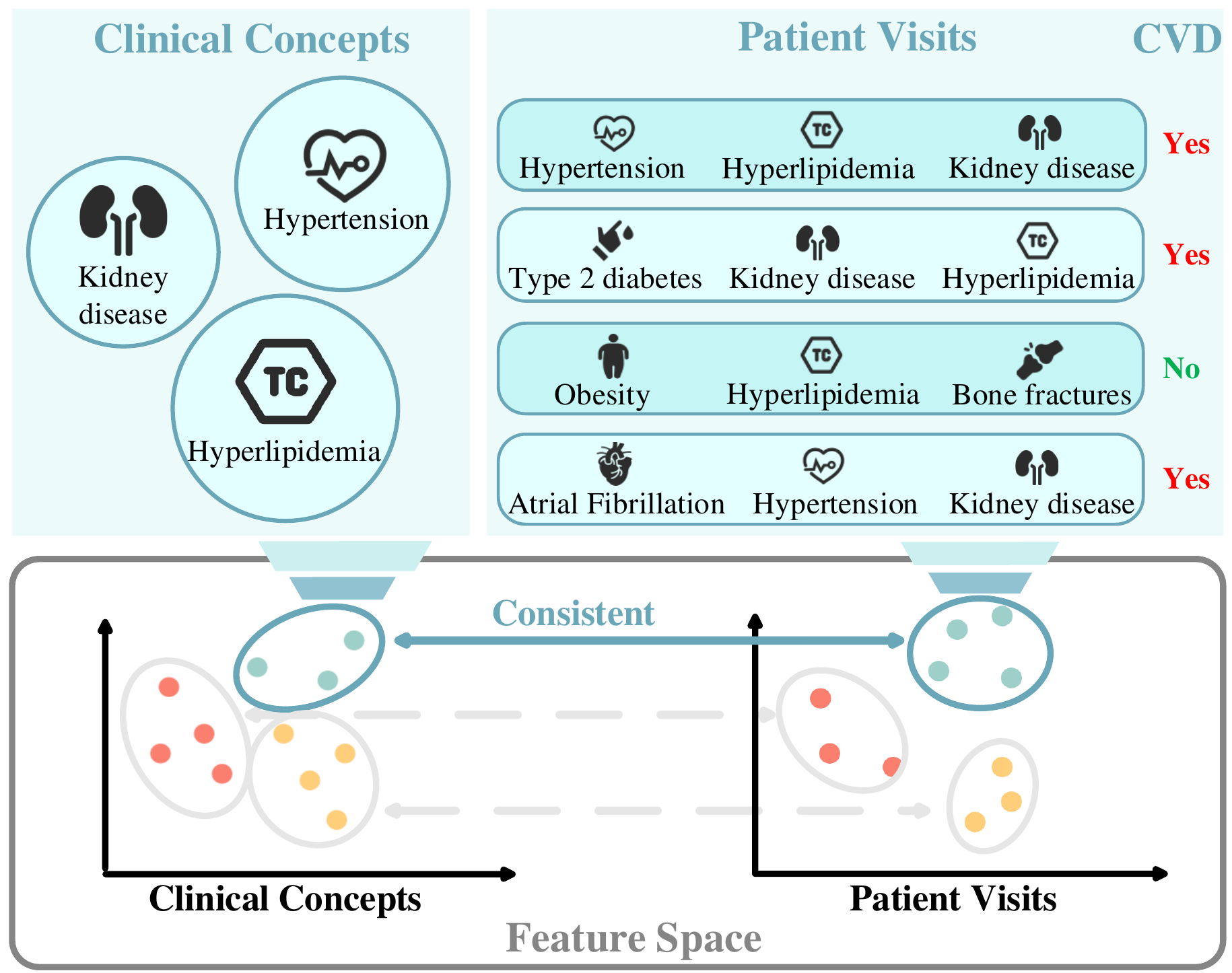}
  \caption{Co-clustering analysis of clinical concepts and patient visits for cardiovascular disease (CVD) identification for diabetic patients. \textnormal{\methodname performs co-clustering over a hypergraph to yield consistent clinical concept and patient visit subgroups. A node cluster of 3 clinical concepts \textit{(blue circle in the left panel)} and a hyperedge cluster of 4 patient visits \textit{(blue circle in the right panel)} suggest potential disease subtypes in correlation with CVD outcomes.}}
 \vspace{0.2cm}
  \label{fig:intro}
\end{figure}

Among various ML models for EHR data, graph-based models have shown promise with their capabilities of modeling complex structures within EHRs \cite{choi2017gram, shang2019pre, liu2020heterogeneous, choi2020learning}. Given the lack of generic ways of constructing reliable graph structures from EHR, hypergraphs have recently been used as a flexible data structure that directly models the interactions between clinical concepts (\textit{i.e.}, names of medical codes in EHR) and patient visits \cite{xu2022counterfactual, xu2023hypergraph}. These higher-order modeling approaches are robust with sparse incomplete features and can generate interpretable predictions on the important clinical concepts for analyzed diseases.  

Understanding disease subtypes is crucial for studying the mechanisms of complex diseases and establishing personalized treatments. Disease subtypes refer to specific variations within a broader disease category, differentiated by unique characteristics, symptoms, or treatment responses. 
In EHR, disease subtypes can be defined as 
subgroups of patients who exhibit similar patterns related to clinical concepts such as diagnoses, medications, and procedures received in their medical visits \cite{wang2023precision, maiorino2023phenomics}.  
In this work, beyond the commonly defined disease types such as in Current Procedural Terminology (CPT) and International Classification of Diseases (ICD), we are interested in further discovering fine-grained disease subtypes such as regarding subgroups of diabetic patients with high risks of stroke, retinopathy, neuropathy, or nephropathy \cite{l2013international}.
On the other hand, modeling disease subtypes and the complex interactions between clinical concepts and patient visits can also assist in the risk prediction of specific diseases. 
For example, in Figure~\ref{fig:intro}, clinical concepts such as hypertension, kidney disease, and hyperlipidemia can indicate a group of patients at high risk of CVD \cite{xu2023hypergraph},
who may require targeted therapies to mitigate the risk of further diabetes-related complications~\cite{ahlqvist2018novel}. 
These connections can help medical professionals gain fine-grained understandings of the risks and design precise effective interventions.

To the best of our knowledge, jointly analyzing the subgroups of clinical concepts and patient visits is rarely studied in healthcare informatics. Previous models have used simple algorithms like K-means \cite{shen2007using, shen2009integrative, ahlqvist2018novel}, single variable analysis \cite{gujral2013type}, and matrix factorization \cite{varghese2021profiles} to identify disease subtypes. However, these statistical methods lack guidance from risk prediction tasks and thus need manually pre-defined sets of clinical concepts for specific diseases \cite{christensen2022type, parikh2023data}. In recent years, more advanced clustering techniques such as Deep Embedded Clustering \cite{xie2016unsupervised} and other self-supervised approaches \cite{zhang2020framework, guo2017improved, li2021contrastive} have been studied for EHR-based clinical predictions. However, these methods can only cluster one type of entity and do not consider the higher-order interactions among clinical concepts and patient visits \cite{huang2021deep, zhao2023subtype, zhang2022modec}.

We aim to develop a model to jointly identify clusters of clinical concepts and patient visits for disease subtyping on EHR data. Task-guided co-clustering is used to analyze clinical concepts and patient visits, providing meaningful interpretations for predictive tasks. However, some challenges need to be addressed:
(1) \textit{Inadequate Graph Representation.} Conventional GNNs struggle to efficiently represent medical concepts and patient visits due to their focus on pairwise relationships. 
EHR data contains multiple medical codes that can be repeated across various visits, requiring a higher-order graph modeling technique for accurate representation. 
Existing work \cite{xu2023hypergraph, ma2018kame, ochoa2022graph, lu2021weighted} only focus on geometric structures and ignore clinical natural language descriptions from medical coding systems, limiting the ability of GNNs to learn a comprehensive representation and negatively impacting downstream tasks.
(2) \textit{Lack of Supervision for Interpretation.} Recent ML research in healthcare has increasingly shifted towards not only showcasing model performance but also providing interpretability. Approaches such as factual counterfactual reasoning \cite{guidotti2019factual, xu2022counterfactual} and time-aware mechanisms \cite{sun2021interpretable, zhang2021context} can extract interpretable subsets for EHR data, but they require supervised learning. In our case, we aim to uncover patterns through co-clustering without supervision and find consistency between clinical concepts and patient visits for interpretations. 

In this work, we introduce \textbf{\methodname} (\underline{\textbf{Ta}}sk-guided \underline{\textbf{C}}\underline{\textbf{o}}-\underline{\textbf{C}}lustering), the first framework that clusters clinical concepts and patient visits on EHR networks using self-supervised co-clustering and a contrastive alignment module. 
Our framework \methodname ``homogenizes'' clusters to reveal deeper insights into the connections between subgroups of clinical concepts and patient visits. 
The contributions of our work are summarized as follows:
\begin{itemize}[nosep,leftmargin=*]
\item We identify a novel task for disease subtyping in EHR analysis, where clusters of clinical concepts and patient visits are jointly studied and contribute to understanding complex diseases. 

\item We develop \methodname, a task-guided self-supervised framework that uncovers patterns through co-clustering without supervision. The model is built based on a text-enhanced hypergraph transformer with a dual application of deep clustering on both nodes and hyperedges. The model further aligns clinical concepts and patient visit clusters through contrastive learning for identifying consistent disease subtypes. 

\item Extensive quantitative experiments and clustering analysis are conducted on two clinical EHR datasets, the publicly available MIMIC-III \cite{johnson2016mimic} and the private CRADLE. 
\methodname outperforms the previous state-of-the-art model \cite{xu2022counterfactual} and demonstrates a notable 5.26\% improvement across four metrics compared to hypergraph model backbone \cite{xu2023hypergraph}. Case studies further show that \methodname is capable of grouping consistent clinical concepts and patient visits that reveal disease subtypes related to a specific disease (e.g., CVD). As validated by a domain expert, the captured disease subtypes could have different levels of relationships (e.g., positive, weak, or negative) with specific diseases for a fine-grained understanding in practice.
\end{itemize}

\vspace{0.2cm}
\section{RELATED WORK}

\textbf{Machine Learning in Healthcare.} 
Medical research has utilized various model architectures to analyze healthcare data. 
Earlier research efforts mainly employed fundamental model architectures. For instance, \citet{liu2014early} used auto-encoders to diagnose Alzheimer’s disease. \citet{choi2016learning} utilized word2vec~\cite{mikolov2013efficient} to learn the representations of medical concepts. More recent models, such as Convolutional Neural Networks and Recurrent Neural Networks have also been widely applied in various applications~\cite{choi2016retain, pham2016deepcare, choi2016doctor, nguyen2016mathtt, harutyunyan2019multitask}.

To capture the intrinsic structures of healthcare data, there is a growing interest in graph-based methods. For example, event sequences are modeled as weight graphs for heart failure prediction~\cite{liu2015temporal}, robust relations among medical codes are learned~\cite{choi2020learning}, and medical knowledge graphs are incorporated for downstream reasoning~\cite{choi2017gram, ma2018kame}. To consider the higher-level relations in structured data, ~\citet{xu2023hypergraph, xu2022counterfactual} proposed to model patient visits as hyperedges in a hypergraph transformer, which overcomes the limitations of pairwise relations and provides interpretable insights.


\vspace{0.2cm}

\noindent\textbf{Clustering Methods.} 
Traditional clustering methods are mostly algorithmic and heuristic-based, e.g., K-means \cite{lloyd1982least}, hierarchical clustering \cite{johnson1967hierarchical}, and density-based spatial clustering of applications with noise (DBSCAN) \cite{ester1996density}. While effective for linearly separable data, these algorithms are not learning-based and cannot generalize to unseen data. To address this limitation, deep clustering techniques like DEC \cite{xie2016unsupervised} introduced self-supervised techniques for complex cluster representation.
Subsequently, several self-supervised clustering approaches \cite{zhang2020framework, guo2017improved, li2021contrastive, yang2016joint, huang2014deep} were proposed to enhance the robustness of clustering and were generalized to various settings.

Clustering techniques have been employed to identify meaningful subsets of medical concepts in healthcare. For example, iCluster \cite{shen2009integrative} jointly estimated the latent tumor subtypes by refining K-means and Gaussian latent variable models. SilHAC \cite{nidheesh2020hierarchical} tackled cluster number estimation and identification in cancer data based on the Silhouette Index. NCIS \cite{liu2014network} identified cancer subtypes based on gene expression with network-assisted co-clustering. MODEC \cite{zhang2022modec} leveraged DEC for cancer subtype identification and clinical feature analysis. However, none of these studies has explored the co-clustering of clinical concepts and patient visits to improve downstream prediction tasks and provide interpretations.

\vspace{0.2cm}
\section{METHOD}
\begin{figure*}[h]
  \centering
\includegraphics[width=0.9\linewidth]{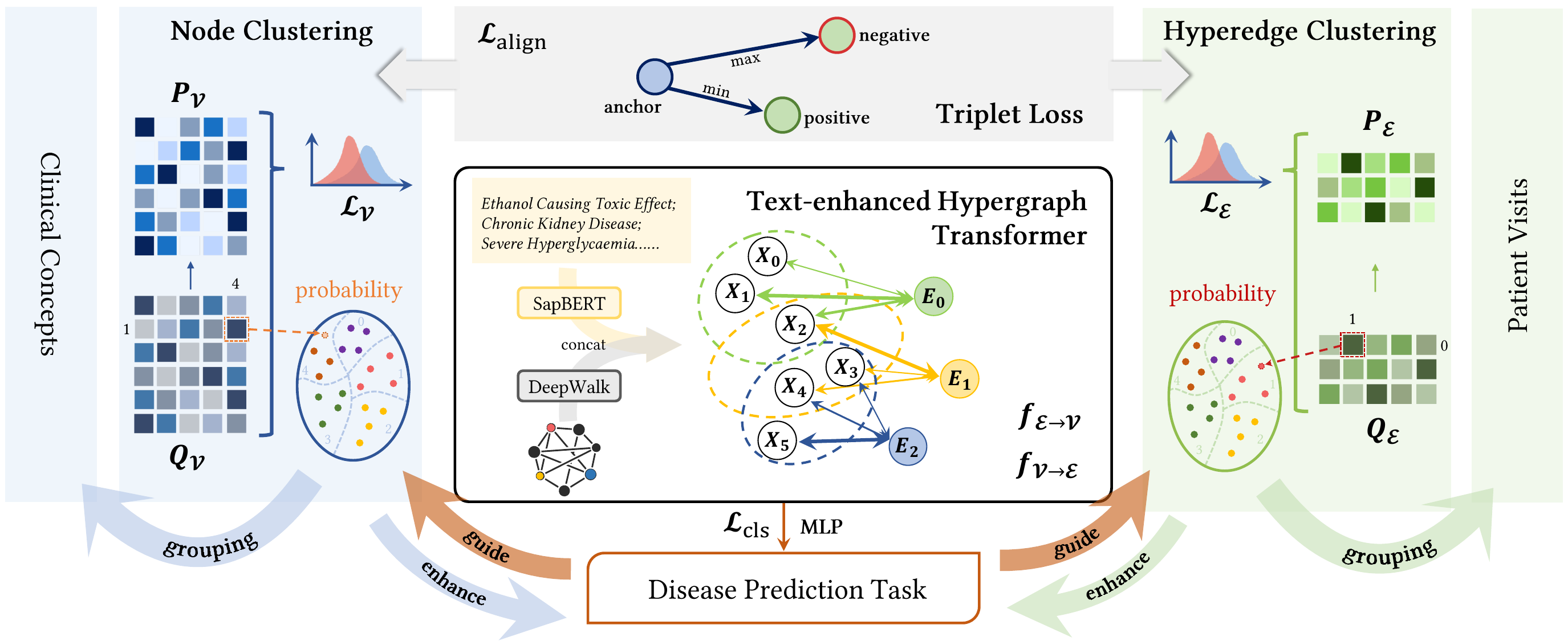}
  \caption{Pipeline of \methodname. \textnormal{A hypergraph transformer \textit{(middle)} is used as a backbone to model node and hyperedge interactions. Node clustering \textit{(left)} and hyperedge clustering \textit{(right)} are jointly optimized to produce clusters of clinical concepts and patient visits. A triplet loss function \textit{(top)} is applied for a consistent cluster alignment across two domains. }}
\vspace{-5pt}
  \label{fig:pipeline}
\end{figure*}

\subsection{Problem Definition}
We define a hypergraph \(\mathcal{G} = (\mathcal{V}, \mathcal{E})\) where \(\mathcal{V}\) is a set of vertices, each representing a medical code, and \(\mathcal{E}\) is a set of hyperedges, each representing a patient visit that includes a subset of medical codes from \(\mathcal{V}\). The goals of this study include: (1) Given a patient's clinical record, predict the clinical outcome \(y\) of that patient. (2)~Analyze clusters of clinical concepts and patient visits for disease subtyping based on the hypergraph \(\mathcal{G}\) constructed from EHR data.

\subsection{Text-enhanced Hypergraph Transformer}
To effectively capture the intricate relationships in EHR, we adopt a hypergraph transformer \cite{chien2021you}. As shown in the middle part of Figure~\ref{fig:pipeline}, the node embedding of each medical code in the hypergraph are initialized with information from two aspects: structures and semantics. The structural part \( \bm{X}_{\text{structure}} \) is obtained from DeepWalk \cite{perozzi2014deepwalk}, where we apply random walks on \(\mathcal{G}\) to train a Skip-gram model. For the semantical part, we process the medical code descriptions of nodes with SapBERT \cite{liu2021self}, a transformer-based model pre-trained on extensive biomedical literature, to generate text embeddings \( \bm{X}_{\text{text}} \). We directly concatenate these two vectors as the initial node embeddings:
\begin{equation}
\setlength{\abovedisplayskip}{5pt}
\setlength{\belowdisplayskip}{5pt}
    \bm{X} = [ \bm{X}_{\text{structure}}; \bm{X}_{\text{text}}].
\end{equation}

For the \( l \)-th layer of the hypergraph, node and hyperedge embeddings are denoted by \( \bm{X}^{(l)} \in \mathbb{R}^{|\mathcal{V}| \times d} \) and \( \bm{E}^{(l)} \in \mathbb{R}^{|\mathcal{E}| \times d'} \), where \( d \) and \( d' \) are dimensionality parameters of the node and hyperedge feature spaces, respectively. The embeddings are updated through a two-step message-passing mechanism:
\begin{equation}
\setlength{\abovedisplayskip}{6pt}
\setlength{\belowdisplayskip}{6pt}
    \bm{E}_e^{(l)} = f_{\mathcal{V} \rightarrow \mathcal{E}}(\mathcal{V}_{e, \bm{X}^{(l-1)}}), \quad \bm{X}_v^{(l)} = f_{\mathcal{E} \rightarrow \mathcal{V}}(\mathcal{E}_{v, \bm{E}^{(l)}}),
    \label{eq:embeddings}
\end{equation}
where \( \mathcal{V}_{e,\bm{X}} = \{\bm{X}_{u,:} : u \in e\} \) is the representation of nodes contained in the hyperedge \( e \), and \( \mathcal{E}_{v,\bm{E}} = \{\bm{E}_{e,:} : v \in e\} \) denote the representations of hyperedges that contain the node \( v \). For the two functions \(f(\cdot)\), we leverage a self-attention mechanism \citep{vaswani2017attention} that allows the model to focus on the most informative parts:
\begin{equation}
\setlength{\abovedisplayskip}{5pt}
\setlength{\belowdisplayskip}{5pt}
\text{Self-Att}(\bm{S}) = \text{LayerNorm}(\bm{Y} + \text{FFN}(\bm{Y})),
\label{eq3}
\end{equation}
where \( \bm{Y} \) is the output from the multi-head self-attention block:
\begin{equation}
\setlength{\abovedisplayskip}{5pt}
\setlength{\belowdisplayskip}{5pt}
\bm{Y} = \text{LayerNorm}(\bm{S} + \big\|_{i=1}^{h} \text{SA}_i(\bm{S})).
\label{eq4} 
\end{equation}
\( \text{SA}_i(\bm{S}) \) denotes the scaled dot-product attention mechanism:
\begin{equation}
\setlength{\abovedisplayskip}{5pt}
\setlength{\belowdisplayskip}{5pt}
\text{SA}_i(\bm{S}) = \text{softmax}\left( \frac{\bm{W}_i^Q(\bm{SW}_i^K)^\top}{\sqrt{\lfloor d/h \rfloor}} \right)\bm{SW}_i^V,
\label{eq:scaled_dot_product}
\end{equation}
where \( \bm{W}_i^Q \), \( \bm{W}_i^K \), and \( \bm{W}_i^V \) are parameter matrices for the \( i \)-th head corresponding to queries, keys, and values, respectively. The input sequence \( \bm{S} \) will be projected into different \( h \) heads. The output of each head is then concatenated (denoted by \( \big\| \)) to form the multi-head attention output. The input dimensionality \( d \) is evenly split across the heads in \(\lfloor d/h \rfloor\) dimensions. The multi-head attention output is combined with a feed-forward neural network (FFN), which is composed of a 2-layer Multilayer Perceptron (MLP) with a ReLU activation. In our task, we do not include the position encoding technique in the standard Transformer due to the lack of such information in our datasets.
A 2-layer MLP is utilized for the disease risk prediction task, along with a sigmoid activation function, denoted by \( \sigma\):
\begin{equation}
\setlength{\abovedisplayskip}{5pt}
\setlength{\belowdisplayskip}{5pt}
\hat{y} = \sigma \left( \text{MLP} \left( \big \|_{l=1}^{L} \hat{\bm{E}}^{(l)} \right) \right).
\label{eqmlp}
\end{equation}
The learning objective is a binary cross-entropy loss, where \(y\) represents the truth label and \(\hat{y}\) is the predicted probability:
\begin{equation}
\setlength{\abovedisplayskip}{5pt}
\setlength{\belowdisplayskip}{8pt}
  \mathcal{L}_{\text{cls}} = -y \log(\hat{y}) - (1 - y) \log(1 - \hat{y}).
  \label{eqloss1}
\end{equation}

\subsection{Deep Self-Supervised Co-clustering}
\label{sec:dec}
The main challenge in our problem is the lack of supervision for generating clusters of clinical concepts and patient visits for disease subtyping. Since there are no labels available, traditional supervised methods such as classification cannot be applied directly. Some previous works use simple K-means to extract specific disease subtypes, but these methods are not data-driven and cannot be integrated into deep models. Inspired by \citet{xie2016unsupervised}, we employ a deep clustering method that iteratively learns the cluster assignments in a self-supervised manner. This deep clustering technique has been proven to be effective in graphs and can be jointly optimized with embedding propagation \cite{yang2020graph}.

In \methodname, we propose a dual application of Deep Embedded Clustering (DEC) on both clinical concepts and patient visits, as shown in the left and right parts of Figure~\ref{fig:pipeline}. Specifically, we seek to jointly learn soft clustering assignments \(\bm{Q}\) for both nodes and hyperedges, denoted as \(\bm{Q}_\mathcal{V}\) and \(\bm{Q}_\mathcal{E}\), respectively. For any given node (or hyperedge) \( i \) and cluster \( k \), the soft assignment \( q_{ik} \) is calculated based on the similarity between the node's (or hyperedge's) embedding and the cluster centroid, formalized as:
\begin{equation}
\setlength{\abovedisplayskip}{5pt}
\setlength{\belowdisplayskip}{5pt}
q_{ik} = \frac{(1 + \| \mathbf{x}_i - \mathbf{u}_k \|^2)^{-1}}{\sum_j (1 + \| \mathbf{x}_i - \mathbf{u}_j \|^2)^{-1}},
\label{eq:soft_assignment}
\end{equation}
where \( \mathbf{x}_i \) is the embedding of node (or hyperedge) \( i \) and \( \mathbf{u}_k \) is the centroid of the \( k \)-th cluster. Similarly to DEC, we perform standard K-means to initialize \(K\) centroids \(\{\mathbf{u}_j\}_{j=1}^K\).

Subsequent to the determination of \( \bm{Q} \), we construct a refined target distribution \( \bm{P} \), which aims to enhance cluster purity by emphasizing confident assignments. The components of \( \bm{P} \) are computed by squaring the elements of \( \bm{Q} \) and normalizing them across each cluster, as follows:
\begin{equation}
\setlength{\abovedisplayskip}{5pt}
\setlength{\belowdisplayskip}{5pt}
p_{ik} = \frac{q_{ik}^2 / f_k}{\sum_j q_{ij}^2 / f_j},
\end{equation}
with \( f_k = \sum_i q_{ik} \) representing the sum of the soft assignments to the \( k \)-th cluster. We then minimize the Kullback-Leibler divergence between \( \bm{Q} \) and \( \bm{P} \), which serves as the self-training clustering loss:
\begin{equation}
\setlength{\abovedisplayskip}{5pt}
\setlength{\belowdisplayskip}{5pt}
\mathcal{L_V} = \mathcal{L_E} = KL(\bm{P}||\bm{Q}) = \sum_i \sum_k p_{ik} \log \frac{p_{ik}}{q_{ik}}.
\end{equation}

\subsection{Cluster Contrastive Alignment}
We further align these clusters across two domains in a shared feature space, which helps to generate consistent clusters on clinical concepts and patient visits for interpretations. The cluster centroids are computed by incorporating the soft assignment probabilities from the matrix \( \bm{Q} \). Specifically, for each cluster \( k \), the centroid \( \mathbf{c}^{k} \) is determined by the weighted average of the node (or hyperedge) embeddings, with the weights given by the soft assignments \( q_{ik} \) for node (or hyperedge) \( i \). This is formally expressed as:
\begin{equation}
\setlength{\abovedisplayskip}{5pt}
\setlength{\belowdisplayskip}{5pt}
\mathbf{c}^{k} = \frac{\sum_{i} q_{ik} \mathbf{x}_i}{\sum_{i} q_{ik}},
\end{equation}
where the denominator \( \sum_{i} q_{ik} \) is the sum of the soft assignments to cluster \( k \). Unlike the rigid nature of hard clustering that strictly assigns nodes/hyperedges based on maximal probability, soft assignment reflects the intrinsic nature of disease subtypes, where clinical concepts (e.g., \textit{obesity}) may serve as potential causes for multiple diseases. With soft assignments, \methodname preserves such natural but often overlooked overlaps present in EHR data.

To enable an unsupervised cluster alignment across two domains, existing strategies \cite{deng2019cluster, wang2022cluster} propose to match the first-order moments of the \(k\)-th cluster from the source domain and target domain. Their loss function is designed to minimize the distance \( \mathcal{D} \) between the node clusters embedding \( \bm{C}_e \in \mathbb{R}^{|\mathcal{E}| \times K} \) and hyperedge clusters embedding \( \bm{C}_v \in \mathbb{R}^{|\mathcal{V}| \times K} \) of the corresponding clusters across two domains. However, these strategies presuppose that the indices of clusters between the two domains are pre-aligned. Our task centers on discovering these correspondences in situations where clusters are randomly generated. Thus, we cannot rely on alignment methods that assume index matching between two domains.

Using the principle of contrastive learning, we align the consistent clusters by minimizing the distance between each node cluster centroid and its nearest hyperedge cluster centroid while maximizing the separation from less similar centroids. This process is illustrated on the top of Figure~\ref{fig:pipeline}. In this design, the node and hyperedge embeddings from Eq.~(\ref{eq:embeddings}) are independently processed by a projection MLP head \cite{grill2020bootstrap}, which contains 2 linear layers with batch normalization and a ReLU activation:
\begin{equation}
\setlength{\abovedisplayskip}{5pt}
\setlength{\belowdisplayskip}{5pt}
\quad \bm{Z}_v = \text{MLP} \left( \bm{C}_v \right), \bm{Z}_e = \text{MLP} \left( \bm{C}_e  \right).
\label{eq:projector}
\end{equation}

We use the triplet loss function to align cross-domain embedding without explicit label correspondences, which is calculated as:
\begin{equation}
\setlength{\abovedisplayskip}{5pt}
\setlength{\belowdisplayskip}{5pt}
\mathcal{L}_\text{align} = \sum_{i=1}^{|\mathcal{V}|} \max(0, \mathcal{D}(\mathbf{z}_v^i, \mathbf{z}_e^{i+}) - \mathcal{D}(\mathbf{z}_v^i, \mathbf{z}_e^{i-}) + m),
\label{eq:align}
\end{equation}
where \(\mathbf{z}_v^i\) (\textit{i.e.}, anchor) is the \(i\)-th embedding from \(\bm{Z}_v\), \(\mathbf{z}_e^{i+}\) is the positive sample for \(\mathbf{z}_v^i\) within \(\bm{Z}_e\), and \(\mathbf{z}_e^{i-}\) represents other negative samples, with \(\mathcal{D}\) measuring distance and \(m\) setting the minimum desired difference between positive and negative samples. In this case, the distance \(\mathcal{D}\) is measured by negative cosine similarity, which has been proven to be effective in contrastive learning \cite{chen2021exploring}:
\begin{equation}
\setlength{\abovedisplayskip}{5pt}
\setlength{\belowdisplayskip}{5pt}
\mathcal{D}(\mathbf{z}_v, \mathbf{z}_e) = -\frac{\mathbf{z}_v \cdot \mathbf{z}_e}{\|\mathbf{z}_v\| \|\mathbf{z}_e\|},
\end{equation}
where \( \cdot \) denotes dot product, and \( \|\cdot\| \) is the \(l_2\)-norm.

\subsection{Learning Objective}
The final learning objective of \methodname is the sum of three parts:
\begin{equation}
\setlength{\abovedisplayskip}{5pt}
\setlength{\belowdisplayskip}{5pt}
\label{eq:loss_all}
    \mathcal{L} = \mathcal{L}_{\text{cls}} + \alpha(\mathcal{L_V} + \mathcal{L_E}) + \beta\mathcal{L}_{\text{align}},
\end{equation}
where \(\alpha\) and \(\beta\) are hyperparameters for weighting different losses.

\vspace{0.2cm}

\definecolor{mypink}{RGB}{242, 121, 112}
\definecolor{mycyan}{RGB}{5, 185, 226}
\definecolor{mypurple}{RGB}{137, 131, 191}

\section{EXPERIMENTS}
In this section, we evaluate \methodname on two EHR datasets, in terms of the performance of downstream clinical tasks, in-depth model analysis, clustering analysis, and case studies.

\subsection{Experiment Settings} \label{chapt:setting}

\noindent\textbf{Datasets and Tasks.}
We adopt the public \textbf{MIMIC-III} dataset~\cite{johnson2016mimic} for a phenotype classification task.  MIMIC-III comprises de-identified health-related data associated with over forty thousand patients who stayed in critical care units of the Beth Israel Deaconess Medical Center between 2001 and 2012. We follow the setting of \cite{harutyunyan2019multitask} to identify 25 phenotypes, including 12 acute conditions (e.g., pneumonia), 8 chronic conditions (e.g., chronic kidney disease), and 5 mixed conditions (e.g., conduction disorders). In terms of experiments, we choose patients with more than one visit and utilize the records from a preceding visit to predict the diagnostic phenotypes of the subsequent visit. These visits are represented as hyperedges within our hypergraph modeling framework, each annotated with a 25-category multihot label.

We also utilize the \textbf{CRADLE} (Emory Clinical Research Analytics Data Lake Environment) dataset for a CVD risk prediction task. Project CRADLE contains close to 48 thousand de-identified patient records with type 2 diabetes seen at Emory Healthcare System between 2013 and 2017. Following \cite{xu2023hypergraph}, our study aims to predict the onset of CVD within one year following the initial diagnosis of type 2 diabetes, utilizing ICD-9 and ICD-10 codes to identify CVD events such as coronary heart disease (CHD), congestive heart failure (CHF), myocardial infarction (MI), or stroke. The patients are considered positive if they develop a CVD complication within a year and negative otherwise.

The dataset statistics are summarized in Table~\ref{tab:stats}.
Our tasks also include a clustering analysis, where task-guided clusters for clinical concepts and patient visits are generated to discover disease subtypes. We also visualize the distribution of disease subtypes and patient subgroups using t-SNE. Their significant relationships are captured and analyzed through case studies in Sec.~\ref{sec:case_study}.

\begin{table}
\centering
\caption{Statistics of MIMIC-III and CRADLE datasets. \textnormal{For MIMIC-III, there are only 12535 are processed with labels.}} 
\vskip -0.5em
\resizebox{0.7\linewidth}{!}{
\begin{tabular}{lcc}
\toprule
\textbf{Stats} & \textbf{MIMIC-III} & \textbf{CRADLE}\\
\midrule
\# of diagnosis & 846 & 7915\\ 
\# of procedure & 2032 & 4321\\ 
\# of service & 20 & ---\\ 
\# of prescription & 4525 & 489\\ 
\# of nodes & 7423 & 12725\\ 
\# of hyperedges & 36875/12353 & 36611\\ 
\bottomrule
\end{tabular}
}
\vspace{0.25cm}
\label{tab:stats}
\end{table}

\vspace{0.11cm}

\noindent\textbf{Metrics.}
To deal with the imbalanced labels within the EHR data, we follow \cite{xu2023hypergraph} to adopt Accuracy, AUROC, AUPR, and Macro F1 score as the metrics of downstream clinical tasks. To measure the quality of clusters \methodname generates, we utilize Silhouette Coefficient \cite{rousseeuw1987silhouettes} for an unsupervised evaluation.

\begin{table*}[!htbp]
\centering
\caption{Performance of clinical outcome predictions on MIMIC-III and CRADLE compared with different baselines. \textnormal{The presented results are averages of the best metrics from 10 individual runs of the models. \textbf{Bold} numbers indicate the best results, and \underline{underlined} numbers indicate the second-best results in each category. TACCO uses \colorbox{gray!20}{DEC} as the default clustering module; we also discuss the performance of DCC and IDEC in our framework. We use * to indicate statistically significant results (\(p \textless  0.05\)).}}
 \vskip -0.5em
\label{tab:model_performance}
\resizebox{\linewidth}{!}{
\begin{tabular}{lcccccccc}
\toprule
\multirow{2.5}{*}{\textbf{Model}} & \multicolumn{4}{c}{\textbf{MIMIC-III}} & \multicolumn{4}{c}{\textbf{CRADLE}} \\
\cmidrule(lr){2-5} \cmidrule(lr){6-9}
 & Accuracy & AUROC & AUPR & Marcro-F1 & Accuracy & AUROC & AUPR & Marcro-F1 \\
\midrule
LR & $68.66 \pm 0.24$ & $64.62 \pm 0.25$ & $45.63 \pm 0.32$ & $13.74 \pm 0.40$ & $76.22 \pm 0.30$ & $57.22 \pm 0.28$ & $25.99 \pm 0.28$ & $42.18 \pm 0.35$ \\
SVM & $72.02 \pm 0.12$ & $55.10 \pm 0.14$ & $34.19 \pm 0.17$ & $32.35 \pm 0.21$ & $68.57 \pm 0.13$ & $53.57 \pm 0.11$ & $23.50 \pm 0.15$ & $52.34 \pm 0.22$ \\
MLP & $70.73 \pm 0.24$ & $71.20 \pm 0.22$ & $52.14 \pm 0.23$ & $16.39 \pm 0.30$ & $77.02 \pm 0.17$ & $63.89 \pm 0.18$ & $33.28 \pm 0.23$ & $45.16 \pm 0.26$ \\
XGBoost & $76.40 \pm 0.42$ & $67.68 \pm 0.35$ & $47.26 \pm 0.34$ & $36.14 \pm 0.59$ & $79.28 \pm 0.26$ & $68.65 \pm 0.58$ & $39.12 \pm 0.39$ & $56.57 \pm 0.65$ \\
\midrule
GCT & $76.58 \pm 0.23$ & $78.62 \pm 0.21$ & $63.99 \pm 0.27$ & $35.48 \pm 0.34$ & $77.26 \pm 0.22$ & $67.08 \pm 0.19$ & $35.90 \pm 0.20$ & $56.66 \pm 0.25$ \\
GAT & $76.75 \pm 0.26$ & $78.89 \pm 0.12$ & $66.22 \pm 0.29$ & $34.88 \pm 0.33$ & $77.82 \pm 0.20$ & $66.55 \pm 0.27$ & $36.06 \pm 0.18$ & $56.43 \pm 0.26$ \\
\midrule
HGNN & $77.93 \pm 0.41$ & $80.12 \pm 0.30$ & $68.38 \pm 0.24$ & $40.04 \pm 0.35$ & $76.77 \pm 0.24$ & $67.21 \pm 0.25$ & $37.93 \pm 0.18$ & $58.05 \pm 0.23$ \\
HyperGCN & $78.01 \pm 0.23$ & $80.34 \pm 0.15$ & $67.68 \pm 0.16$ & $39.29 \pm 0.20$ & $78.18 \pm 0.11$ & $67.83 \pm 0.18$ & $38.28 \pm 0.19$ & $60.24 \pm 0.21$ \\
HCHA & $78.07 \pm 0.28$ & $80.42 \pm 0.17$ & $68.56 \pm 0.15$ & $37.78 \pm 0.22$ & $78.60 \pm 0.15$ & $68.05 \pm 0.17$ & $39.23 \pm 0.13$ & $59.26 \pm 0.21$ \\
HypEHR & {$79.07 \pm 0.31$} & {$82.19 \pm 0.13$} & {$71.08 \pm 0.17 $} & {$41.51 \pm 0.25$} & {$79.76 \pm 0.18$} & {$70.07 \pm 0.13$} & {$40.92 \pm 0.12$} & {$61.23 \pm 0.18$} \\
\midrule
\textbf{TACCO} & \multicolumn{8}{c}{} \\
\quad w/ DCC & $79.56 \pm 0.25^*$ & $82.47 \pm 0.15^*$ & $71.37 \pm 0.29^*$ & $40.45 \pm 0.62^*$ & \underline{$80.24 \pm 0.30^*$} & $72.67 \pm 0.23^*$ & $45.48 \pm 0.44^*$ & $61.17 \pm 0.54^*$\\
\quad w/ IDEC & \underline{$80.75 \pm 0.09^*$} & \underline{$84.08 \pm 0.22^*$} & \underline{$73.63 \pm 0.28^*$} & \bm{$45.59 \pm 0.67^*$} & $80.06 \pm 0.23^*$ & \underline{$73.48 \pm 0.26^*$} & \underline{$48.09 \pm 0.45^*$} & \underline{$64.55 \pm 0.15^*$} \\
\rowcolor{gray!20} \quad w/ DEC & \bm{$81.02 \pm 0.26^*$} & \bm{$84.31 \pm 0.15^*$} & \bm{$73.67 \pm 0.25^*$} & \underline{$45.53 \pm 0.17^*$} & \bm{$81.00 \pm 0.32^*$} & \bm{$74.23 \pm 0.36^*$} & \bm{$49.08 \pm 0.43^*$} & \bm{$64.64 \pm 0.57^*$} \\
\bottomrule
\end{tabular}
}
\end{table*}

\begin{table*}
\centering
\vskip 0.1em
\caption{Ablation studies on MIMIC-III and CRADLE. \textnormal{The presented results are averages of the best metrics from 10 individual runs of the models. All models are based on the same backbone model hypergraph transformer. \textit{text} refers to the use of textual information, \textit{node} means clustering on nodes, \textit{edge} means clustering on hyperedges, and \textit{align} represents cluster alignment.}}
 \vskip -0.5em
\label{tab:ablation}
\resizebox{\linewidth}{!}{
\begin{tabular}{cccc|cccccccc}
\toprule
\multicolumn{4}{c}{\textbf{hypergraph w/}} & \multicolumn{4}{c}{\textbf{MIMIC-III}} & \multicolumn{4}{c}{\textbf{CRADLE}} \\
\cmidrule(lr){1-4} \cmidrule(lr){5-8} \cmidrule(lr){9-12}
text & node & edge & align & Accuracy & AUROC & AUPR & Macro-F1 & Accuracy & AUROC & AUPR & Macro-F1 \\
\midrule
 &  &  &  & $79.07 \pm 0.31$ & $82.19 \pm 0.13$ & $71.08 \pm 0.17 $ & $41.51 \pm 0.25$ & $79.76 \pm 0.18$ & $70.07 \pm 0.13$ & $40.92 \pm 0.12$ & $61.23 \pm 0.18$ \\
\checkmark &  &  &  & $80.69 \pm 0.28$ & $83.71 \pm 0.38$  & $72.96 \pm 0.30$ & \underline{$45.50 \pm 0.41$} & $80.30 \pm 0.41$ & $73.47 \pm 0.22$ & $47.66 \pm 0.28$ & $64.39 \pm 0.59$ \\
\checkmark & \checkmark &  &  & $80.66 \pm 0.06$ & $83.97 \pm 0.08$ & $72.96  \pm 0.12$ & $45.20 \pm 0.54$ & $80.55 \pm 0.22$ & $73.60 \pm 0.19$ & $47.67 \pm 0.49$ & $63.94 \pm 0.69$ \\
\checkmark &  & \checkmark &  & $80.73 \pm 0.05$ & $84.04 \pm 0.08$ & $73.10 \pm 0.12$ & $45.01 \pm 0.33$ & $80.55 \pm 0.20$ & \underline{$73.82 \pm 0.19$} & $47.75 \pm 0.32$ & \underline{$64.45 \pm 0.59$} \\
\checkmark & \checkmark & \checkmark &  & \underline{$80.77 \pm 0.08$} & \underline{$84.10 \pm 0.06$} & \underline{$73.52 \pm 0.16$} & $45.21 \pm 0.47$ & \underline{$80.73 \pm 0.26$} & $73.73 \pm 0.25$ & \underline{$48.19 \pm 0.58$} & $64.27 \pm 0.68$ \\
\checkmark & \checkmark & \checkmark & \checkmark & \bm{$81.02 \pm 0.26$} & \bm{$84.31 \pm 0.15$} & \bm{$73.67 \pm 0.25$} & \bm{$45.53 \pm 0.17$} & \bm{$81.00 \pm 0.32$} & \bm{$74.23 \pm 0.36$} & \bm{$49.08 \pm 0.43$} & \bm{$64.64 \pm 0.57$} \\
\bottomrule
\end{tabular}
}
\end{table*}

\vspace{0.25cm}

\noindent\textbf{Baselines.} In terms of the two disease prediction tasks, we compare our \methodname with the following baselines:

\noindent  \textit{(1) Traditional ML.} The traditional ML baselines include models that process EHR data without utilizing graph structures. Specifically, we consider:
\begin{itemize}[nosep,leftmargin=*]
    \item \textbf{LR \cite{menard2002applied}}: Logistic Regression, a linear model for classification that estimates probabilities using a logistic function.
    \item \textbf{SVM \cite{cortes1995support}}: Support Vector Machine, an algorithm that finds the hyperplane that best separates different classes.
    \item \textbf{MLP \cite{naraei2016application}}: Multilayer Perceptron, an artificial neural network that consists of at least three layers of nodes.
    \item \textbf{XGBoost \cite{chen2016xgboost}}: an implementation of gradient-boosted decision trees designed for speed and performance.
\end{itemize}

\noindent  \textit{(2) Graph-based Models.} We also compare advanced GNN models for further analysis of EHR data. In this category, we evaluate:
    \begin{itemize}[nosep,leftmargin=*]
        \item \textbf{GCT \cite{choi2020learning}}: Graph Convolutional Transformer, an improved GNN that combines mechanisms of convolution with attention.
        \item \textbf{GAT \cite{velivckovic2017graph}}: Graph Attention Network, an improved GNN that uses attentions mechanism to weight the significance of nodes.
    \end{itemize}
    
\noindent \textit{(3) Hypergraph-based Models.} Hypergraph modeling has become the state-of-the-art approach in EHR analysis. We select several representative methods including:
    \begin{itemize}[nosep,leftmargin=*]
        \item \textbf{HGNN \cite{feng2019hypergraph}}: Hypergraph Neural Network, a hypergraph model that learns the hidden representation via high-order structures.
        \item \textbf{HyperGCN \cite{yadati2019hypergcn}}: Hypergraph Convolutional Network, a model that uses convolution for semi-supervised learning based on higher-order graph modeling.
        \item \textbf{HCHA \cite{bai2021hypergraph}}: Hypergraph Convolution and Hypergraph Attention, a hypergraph model that integrates both convolution and attention mechanisms.
        \item \textbf{HypEHR \cite{xu2023hypergraph}}: A hypergraph transformer based on AllSetTransformer \cite{chien2021you} that predicts disease risks on EHR data.
    \end{itemize}

\vspace{0.1cm}
We also compare several clustering methods with the default DEC in \methodname in terms of the clustering quality:

    \begin{itemize}[nosep,leftmargin=*]
        \item \textbf{HDBSCAN \cite{mcinnes2017accelerated}}: Hierarchical Density-Based Spatial Clustering of Applications with Noise, a clustering algorithm that identifies clusters of varying densities by building a hierarchy of clusters using a density-based approach.
        \item \textbf{IDEC \cite{guo2017improved}}: Improved Deep Embedded Clustering, which improves DEC by introducing an additional autoencoder for embedding reconstructions.
        \item \textbf{DCC \cite{zhang2020framework}}: Deep Constraint Clustering, which explores different constraints that benefit the clustering performance. In our implementation, we choose a global size constraint that assumes each cluster should be approximately the same size.
    \end{itemize}

\vspace{0.2cm}

\noindent\textbf{Implementation Details.}
All the experiments are run on one NVIDIA H100 Tensor Core GPU. We implement most of our experiments using PyTorch\footnote{https://pytorch.org/}. For traditional ML baselines, we implement the code with scikit-learn\footnote{https://scikit-learn.org/}. For the other baselines, we follow the original settings suggested by the authors to train them. For our \methodname, we use Adam optimizers for all modules and tune the learning rates in \(\{5e-4, 1e-3, 5e-3, 1e-2\}\). The frozen SapBERT encoder is implemented from Hugging Face\footnote{https://huggingface.co/cambridgeltl/SapBERT-from-PubMedBERT-fulltext}. To ensure a fair comparison, we strictly adhere to the hyperparameter settings of the backbone hypergraph \cite{xu2023hypergraph}, with \(L=3\) layers, \(h=4\) heads, and \(d=48\) as the hidden feature dimension in each hypergraph layer. For our best model in Table~\ref{tab:model_performance}, we set \(\alpha=10\), \(\beta=0.1\), \(K=5\), and \(m=1\).  The datasets are split into train/validation/test sets by 7:1:2. For the training, we warm up the hypergraph transformer alone with 100 epochs. For further insights into our hyperparameter selection, please refer to Sec.~\ref{chapt:hyperparameter}.

\vspace{0.2cm}
\subsection{Overall Performance Comparison}
We perform two downstream tasks, \textit{i.e.}, the phenotype classification on MIMIC-III and the CVD risk prediction on CRADLE, to evaluate the predictive performance of our proposed \methodname. The comparison with other baselines we discuss in Sec.~\ref{chapt:setting} is presented in Table~\ref{tab:model_performance}.
It is evident that \methodname  consistently outperforms all baselines on four metrics across both datasets, with DEC performing the best as the clustering module. We can observe a significant improvement of 31.25\% on top of the traditional ML models, which often suffer from the sparse nature of large-scale EHR networks. \methodname also gains an improvement of 12.36\% over the two graph-based models. These notable improvements highlight the effectiveness of modeling higher-order relations~within~complex~EHR~data. 

Hypergraph-based models such as HCHA demonstrate better results compared to the traditional ones as they model EHR data beyond pairwise relations and learn robust representations. Compared to these advanced approaches, \methodname still maintains its lead by an average improvement of 7.89\% across two datasets. Compared with the suboptimal HypEHR, our model raises the overall performance by 5.26\%. These results further validate the effectiveness of our task-guided co-clustering in terms of improving downstream~predicting.

\vspace{0.2cm}
\subsection{In-depth Model Analysis}

\noindent\textbf{Model Ablation.}
We conduct detailed ablation studies to better understand the efficacy of different components in \methodname. As shown in Table~\ref{tab:ablation}, we observe a significant improvement in all four metrics when semantic information is involved compared to the backbone model. This suggests that clinical concept semantics are vital in providing more information for structured modeling. Additionally, both node and hyperedge clustering techniques improve the model's predictive power, with hyperedge clustering bringing a slightly better gain. The model's performance is further improved by equipping it with cross-domain alignment loss, which minimizes the distance between similar cluster centroids to generate more consistent results. Optimal performance is achieved when all the clustering and aligning components are integrated. 
This highlights the collective contribution of all four proposed components towards the enhanced model performance, with better interpretability as an additional benefit as shown in Sec.~\ref{sec:case_study}.

\begin{figure}[h]
    \centering
    \vspace{0.2cm}
    \subcaptionbox{Varying \(\alpha\)}[.33\linewidth]{
        \includegraphics[width=\linewidth]{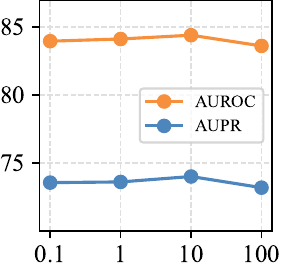}
        \vspace{-4ex}} 
    \subcaptionbox{Varying \(\beta\)}[.326\linewidth]{
        \includegraphics[width=\linewidth]{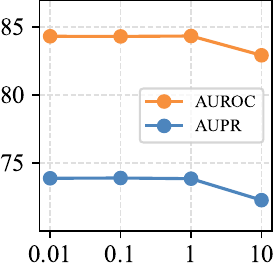}
        \vspace{-4ex}} 
    \subcaptionbox{Varying \(K\)}[.326\linewidth]{
        \includegraphics[width=\linewidth]{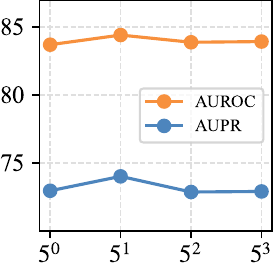}
        \vspace{-4ex}} 
    \vskip 0.5em
    \subcaptionbox{Varying \(\alpha\)}[.33\linewidth]{
        \includegraphics[width=\linewidth]{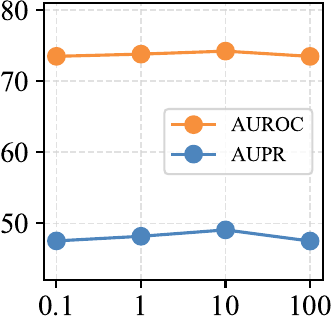}
        \vspace{-4ex}} 
    \subcaptionbox{Varying \(\beta\)}[.326\linewidth]{
        \includegraphics[width=\linewidth]{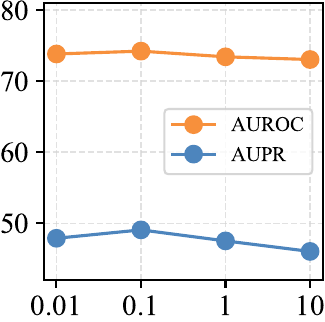}
        \vspace{-4ex}}
    \subcaptionbox{Varying \(K\)}[.326\linewidth]{
        \includegraphics[width=\linewidth]{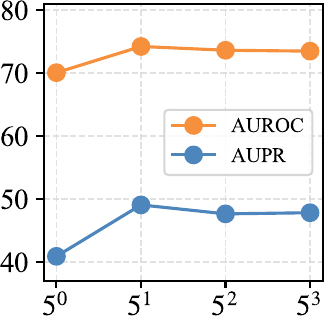}
        \vspace{-4ex}}

    \caption{Effect of hyperparameters of \methodname. \textnormal{(a), (b), and (c) are on MIMIC-III dataset. (d), (e), and (f) are on CRADLE dataset.}}
 
     \label{fig:hyper}
\end{figure}

\noindent\textbf{Hyperparameter Study.} \label{chapt:hyperparameter}
We analyze the impact of important hyperparameters in our \methodname model, which includes the loss weight parameters $\alpha$ and $\beta$ in Eq.~(\ref{eq:loss_all}), and the number of clusters $K$. We vary the contribution of different terms in Eq.~(\ref{eq:loss_all}) by adjusting their respective loss weights, as their numerical values are at different scales. The results are displayed in Figure~\ref{fig:hyper}. Our findings indicate that the best performance is attained when $\alpha$ is set at 10 and $\beta$ is set at 0.1. We also change the number of clusters $K$ and select the value that yielded the best performance for the model.

\begin{figure}[t]
  \centering
  \includegraphics[width=0.8\linewidth]{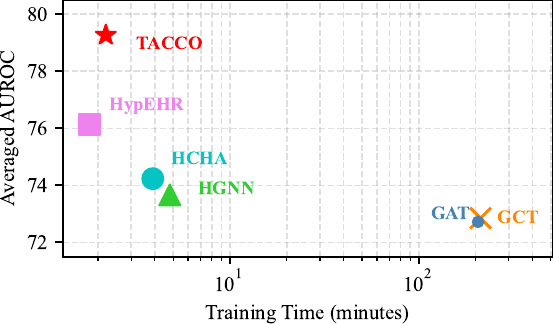}
  \vspace{-4pt}
  \caption{Efficiency and performance trade-off. \textnormal{The y-axis represents the model performance, measured by the averaged AUROC scores from the MIMIC-III and CRADLE datasets. The x-axis shows the training time using a logarithmic scale, also averaged from the two datasets. Some traditional ML methods are not included due to the scikit-learn package's inability to utilize GPU acceleration.}}
 \vspace{0.2cm}
  \label{fig:eff}
\end{figure}

\vspace{0.15cm}
\noindent\textbf{Efficiency Study.} \label{chapt:eff}
Efficiency experiments on our downstream tasks reveal that \methodname achieves the best trade-off of efficiency and performance. As shown in Figure~\ref{fig:eff}, while \methodname's training time is slightly longer than HypEHR's by less than a minute, it achieves the highest AUROC scores in both MIMIC-III and CRADLE datasets. \methodname also easily outperforms other graph-based and hypergraph-based models. Especially for GCT and GAT, their extended training times can be attributed to their lack of hypergraph architecture, which results in significant computational overhead due to the need to flatten all hyperedges.


\begin{table}
\centering
\caption{Clustering quality comparison. \textnormal{SC denotes the Silhouette Coefficient. $\mathcal{V}$ and $\mathcal{E}$ denote the metrics that are calculated in node clustering and hyperedge clustering, respectively. Each method is executed with random seeds from 1 to 5 to ensure the stability of learned embeddings. The results are averaged.}}
 \vskip -0.5em
\label{tab:in_domain}
\resizebox{0.8\linewidth}{!}{
\begin{tabular}{lcccc}
\toprule
\multirow{2.5}{*}{\textbf{Model}} & \multicolumn{2}{c}{\textbf{MIMIC-III}} & \multicolumn{2}{c}{\textbf{CRADLE}} \\
\cmidrule(lr){2-3} \cmidrule(lr){4-5}
 & $\text{SC}_\mathcal{V}$ & $\text{SC}_\mathcal{E}$ & $\text{SC}_\mathcal{V}$ & $\text{SC}_\mathcal{E}$ \\
\midrule
HDBSCAN & 0.1270 & 0.4186 & 0.0399 & 0.0461 \\
K-means ($K=5$) & 0.4131 & \underline{0.8811} & 0.1715 & 0.2313 \\
K-means ($K=10$) & 0.2852  & 0.6350  & 0.1136 & 0.1531 \\
\midrule
\textbf{TACCO ($K=5$)} & \multicolumn{4}{c}{} \\
\quad w/ DCC & 0.0959 & 0.3476 & 0.1390 & 0.1441 \\
\quad w/ IDEC & 0.4993 & 0.7639 & 0.2197 & 0.2226 \\
\rowcolor{gray!20} \quad w/ DEC & \underline{0.4999}  & \textbf{0.8881}  & \textbf{0.3033} & \textbf{0.5083} \\
\midrule
\textbf{TACCO ($K=10$)} & \multicolumn{4}{c}{} \\
\quad w/ DCC & 0.1209 & 0.2834 & 0.1717 & 0.0879 \\
\quad w/ IDEC & 0.4877 & 0.7648 & 0.2044 & 0.2204 \\
\rowcolor{gray!20} \quad w/ DEC & \textbf{0.5849}  & 0.7888  & \underline{0.2242} & \underline{0.3727} \\

\bottomrule
\end{tabular}
}
\vskip 1em
\end{table}

\begin{figure}[!h]
    \centering
    \subcaptionbox{\(K\)=5, node}[.49\linewidth]{
        \includegraphics[width=\linewidth]{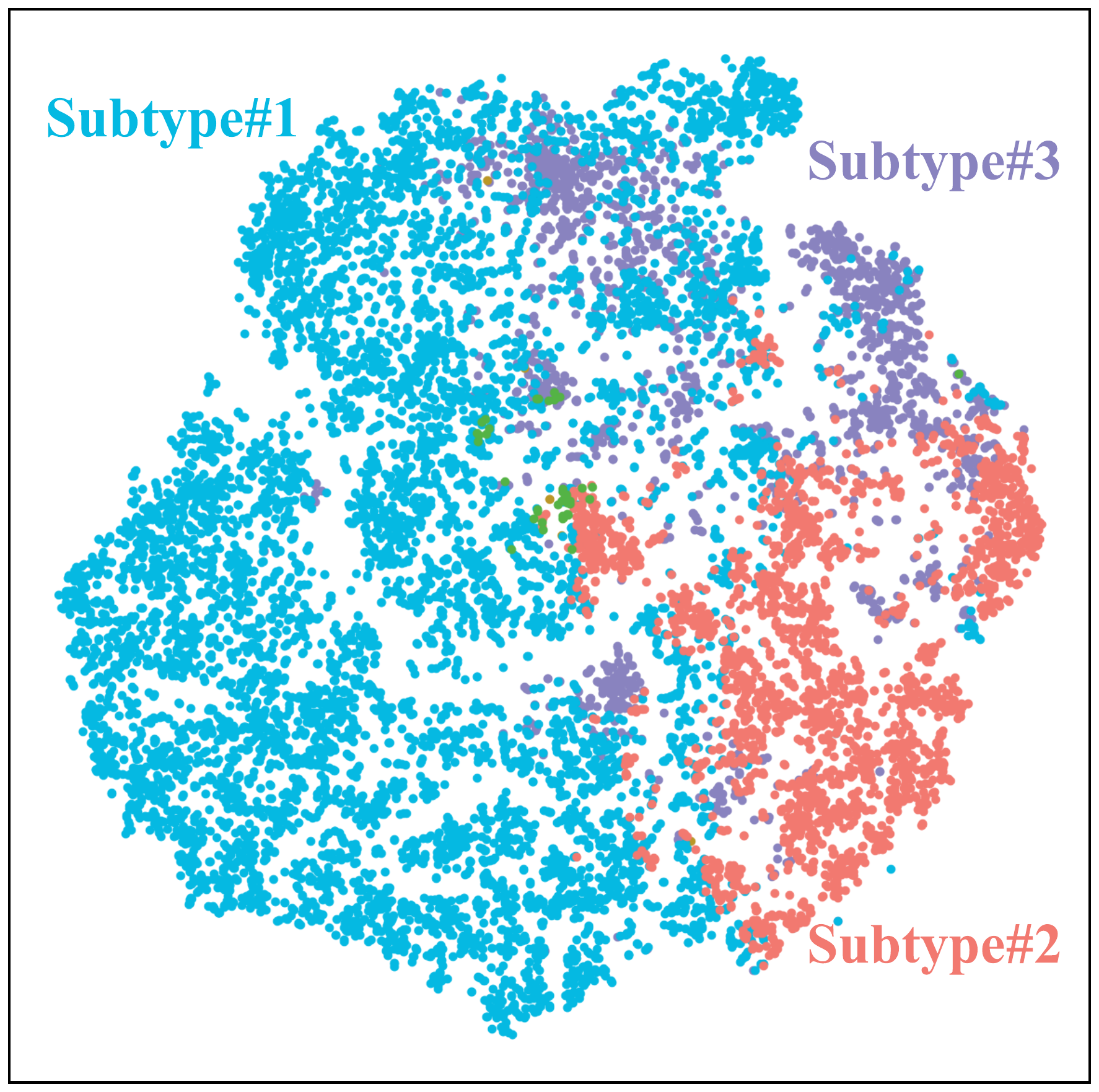}\label{fig:node_k5}}
    \subcaptionbox{\(K\)=10, node}[.49\linewidth]{
        \includegraphics[width=\linewidth]{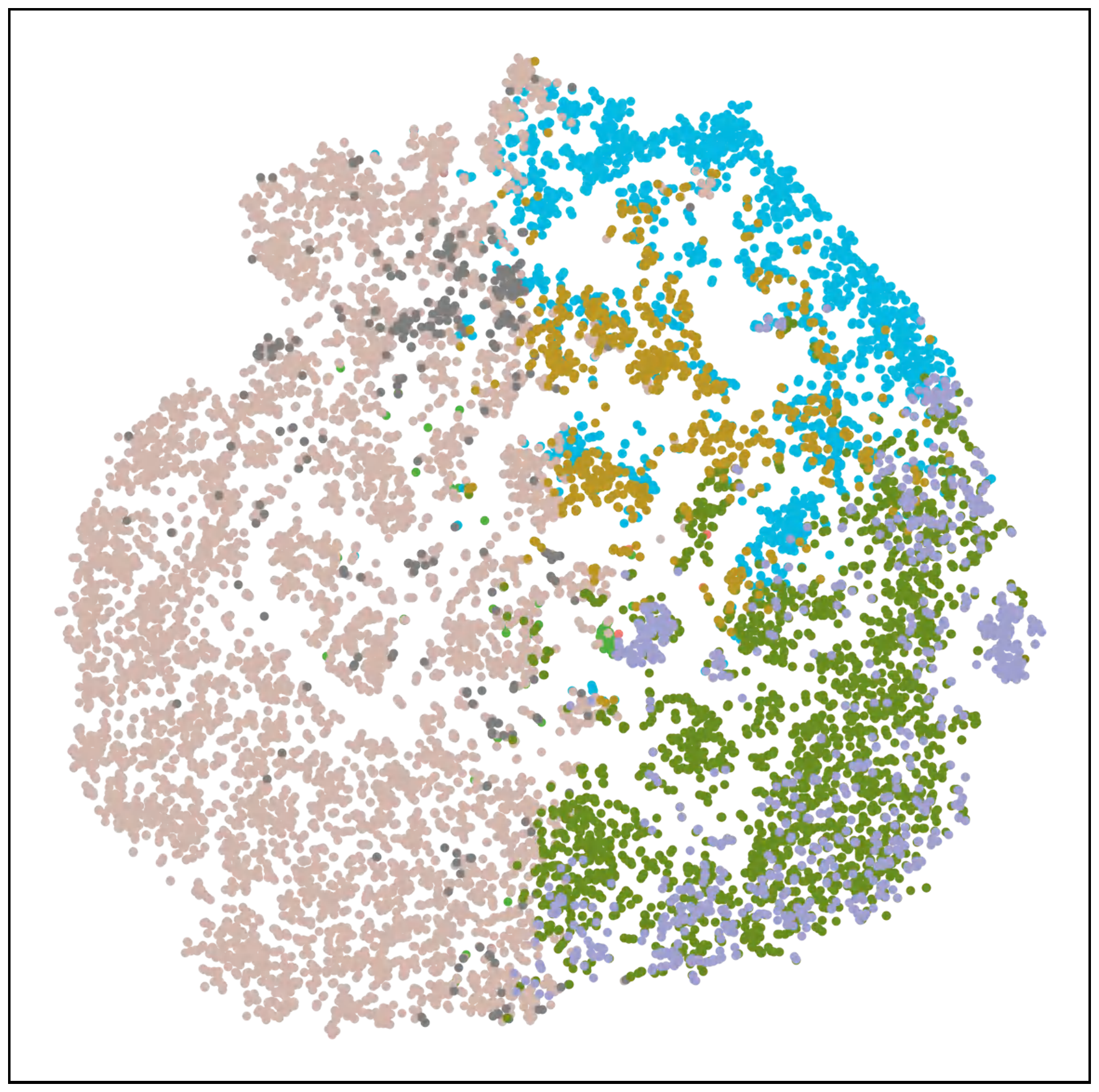}\label{fig:node_k10}}
    \subcaptionbox{\(K\)=5, hyperedge}[.49\linewidth]{
        \includegraphics[width=\linewidth]{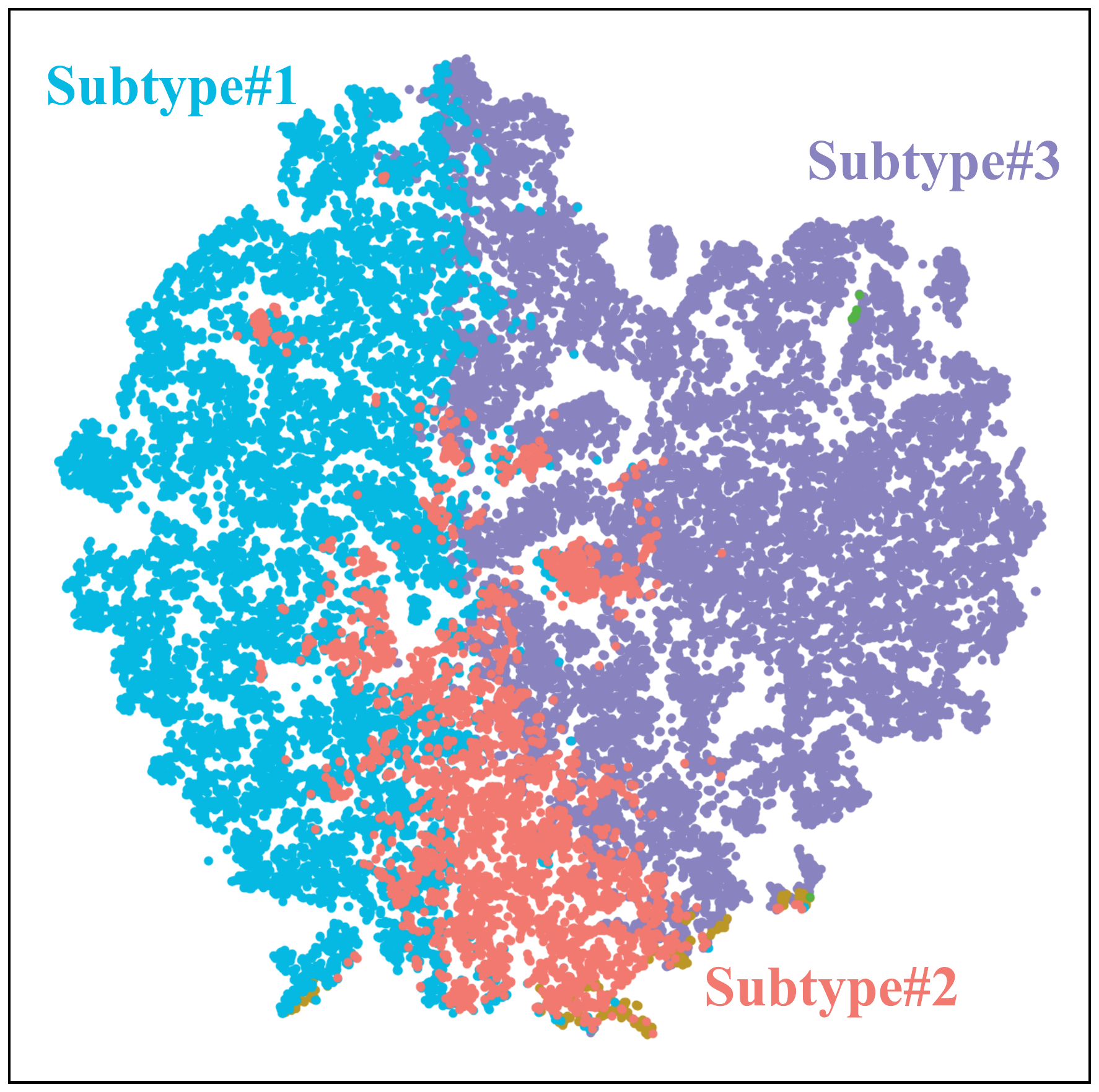}\label{fig:edge_k5}}
    \subcaptionbox{\(K\)=10, hyperedge}[.49\linewidth]{
        \includegraphics[width=\linewidth]{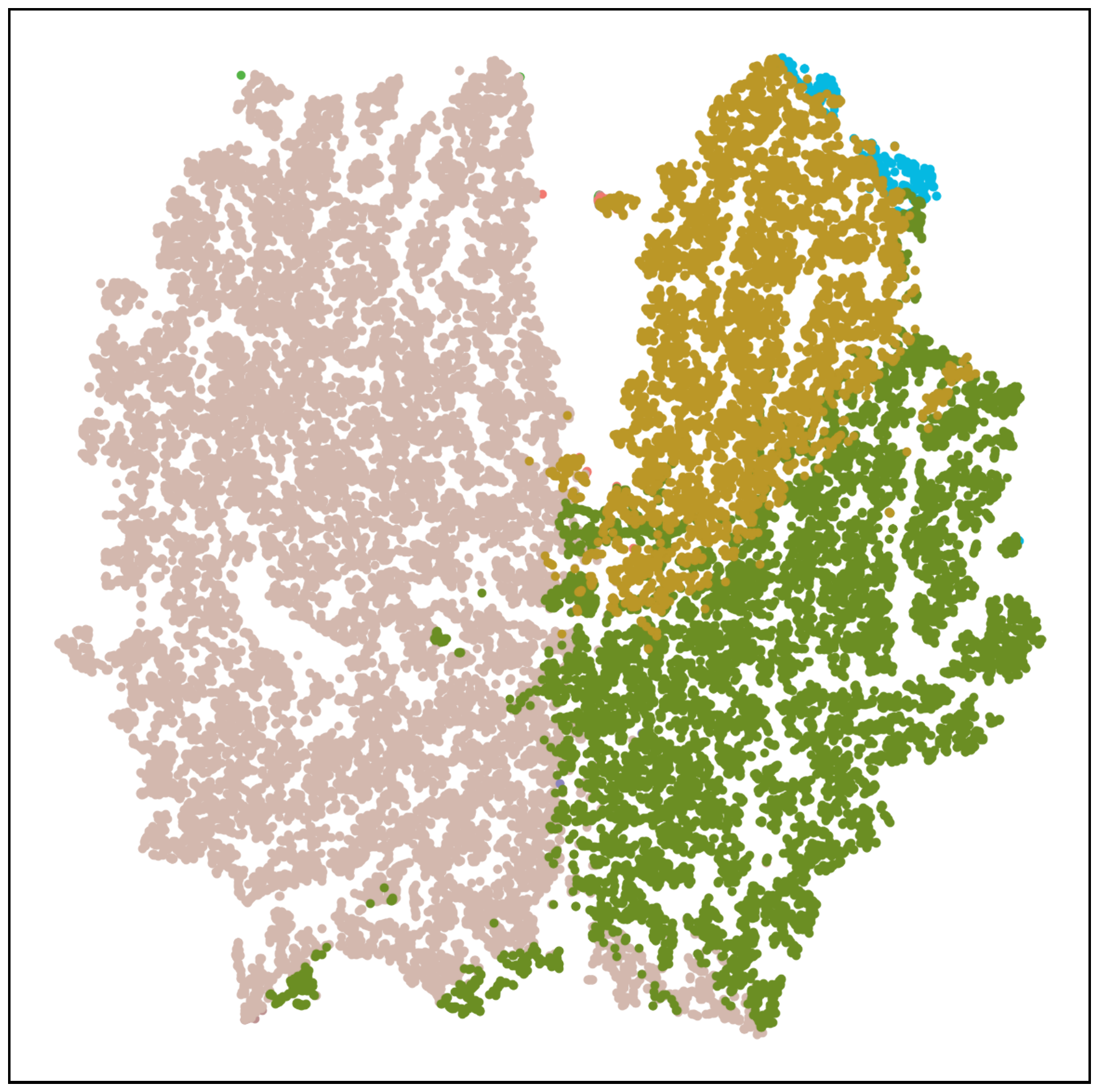}\label{fig:edge_k10}}
    \subcaptionbox{\(K\)=5, CVD Ground Truth}[.49\linewidth]{
        \includegraphics[width=\linewidth]{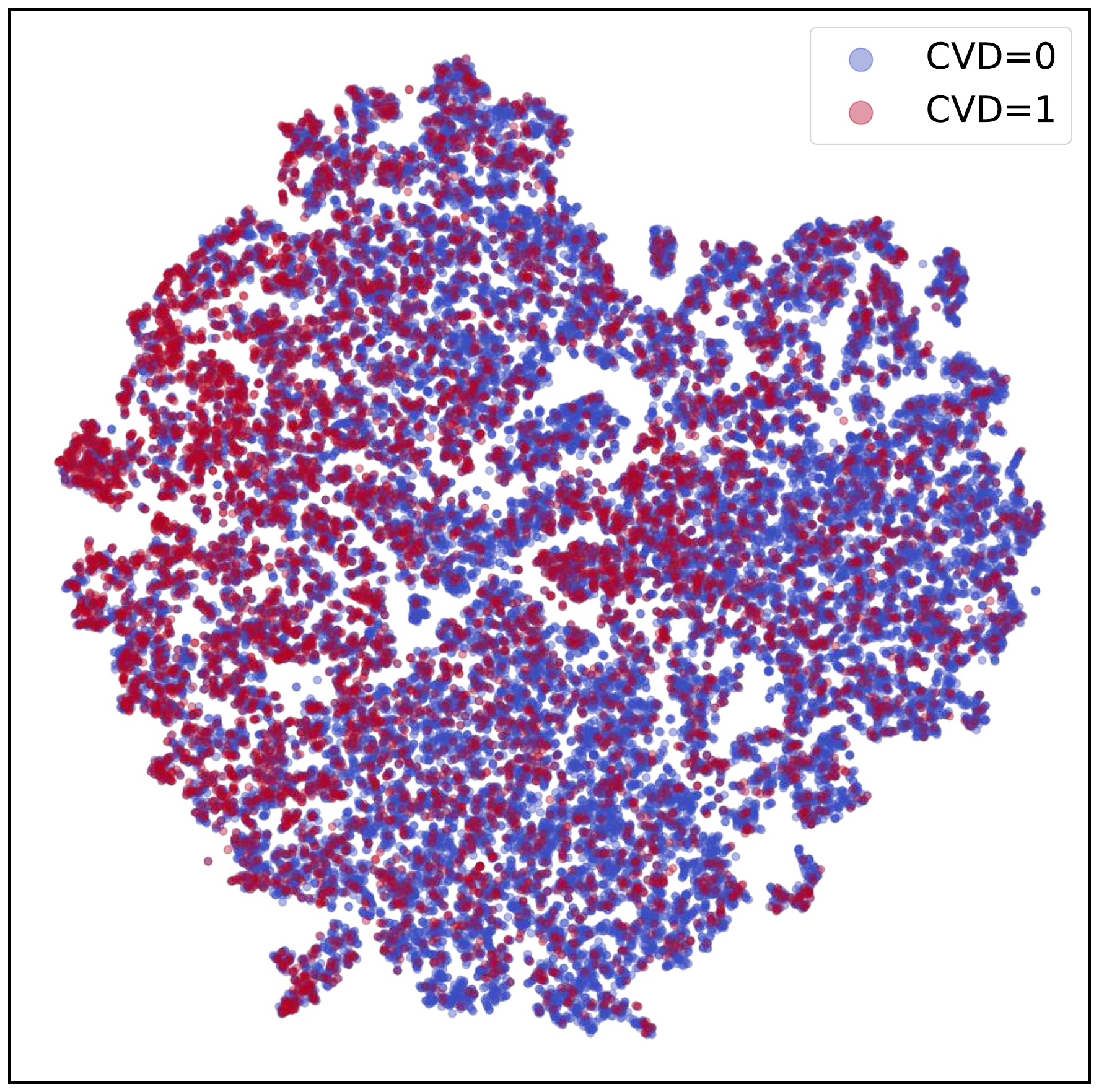}\label{fig:GT_k5}}
    \subcaptionbox{\(K\)=10, CVD Ground Truth}[.49\linewidth]{
        \includegraphics[width=\linewidth]{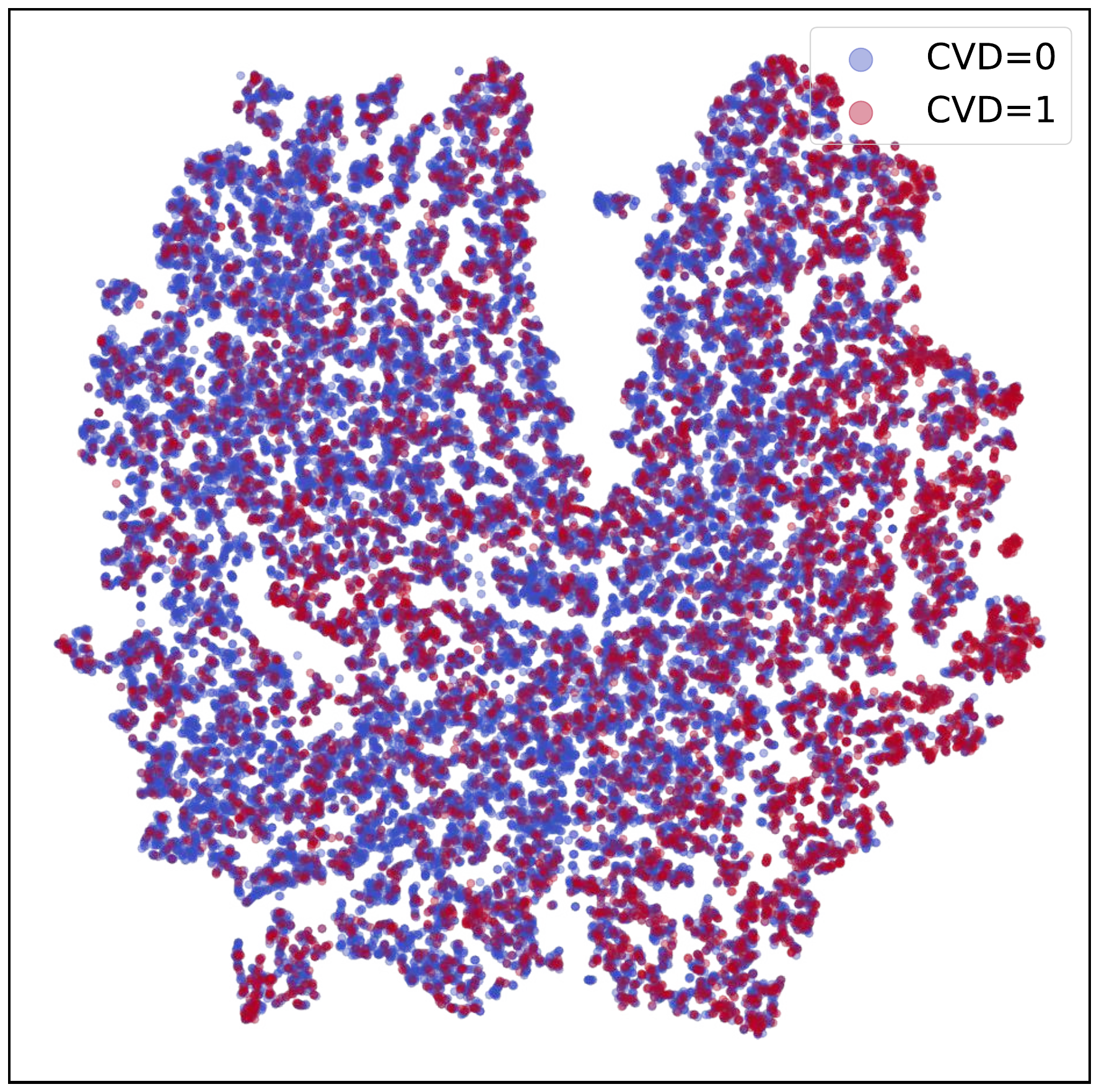}\label{fig:GT_k10}} 

    \caption{Visualization of output clustering distribution and CVD ground truth on the CRADLE dataset via t-SNE. \textnormal{(a) and (c) show clusters distribution on nodes and hyperedges, (e) labels CVD on hyperedges embeddings, all from a \methodname with \(K=5\). (b), (d), and (f) are in the same order from a model with \(K=10\).}}

     \vskip 0.5em
    \label{fig:clus}
\end{figure}

\vspace{0.2cm}
\subsection{Clustering Analysis}
\label{sec:clus_an}
To investigate the quality of our co-clustering assignment, we run the model on the CRADLE dataset with \(K=5\) and \(K=10\), respectively. The high-dimensional embeddings of nodes and hyperedges are then projected into a shared 2D space via t-SNE \cite{van2008visualizing}. The visualizations are presented in Figure~\ref{fig:clus}, with (a), (c), and (e) from the first model with \(K=5\), as well as (b), (d), and (f) from the second model with \(K=10\). The clustering outcomes reveal that, regardless of whether 5 or 10 clusters are targeted, the results consistently form 3-4 major distinct clusters. This consistency highlights the model's stable ability to capture the major patterns underlying the interactions of clinical concepts and patient visits within EHR data. Notably, the clusters on hyperedges and ground-truth CVD labels on hyperedges demonstrate a certain level of concordance. When \(K=5\), the \textcolor{mycyan}{cyan} cluster in panel (c) (denoted as \textcolor{mycyan}{Subtype\#1}) largely coincides with the population  diagnosed with CVD. This indicates that our self-supervised clustering on hyperedges is significantly guided by signals from the specific downstream disease prediction task. Such guiding signals are further propagated to the node clusters through our contrastive alignment objective in Eq.~(\ref{eq:align}), as evidenced by the similarity in color distributions observed between panels (a) and (c), as well as (b) and (d). 

From a more quantitative perspective, we provide a clustering comparison on MIMIC-III and CRADLE datasets in Table~\ref{tab:in_domain} with HDBSCAN \cite{mcinnes2017accelerated} and K-means applied post-training at the final epoch of the hypergraph model. We also discuss the how different deep clustering methods (IDEC \cite{guo2017improved} and DCC \cite{zhang2020framework}) perform in our framework. We take the Silhouette Coefficient as our metric, which offers a robust measure of cluster purity and separation. The Silhouette Coefficient is defined as \(s(i) = \frac{b(i) - a(i)}{\max\{a(i), b(i)\}}\), where \(a(i)\) is the average intra-cluster distance, \textit{i.e.}, cohesion, and \(b(i)\) is the average distance to the nearest cluster that \(i\) is not part of, \textit{i.e.}, separation. We can observe that \methodname using the default DEC (Sec.~\ref{sec:dec}) generates clusters with the highest quality in both nodes and hyperedges. The other deep clustering methods show suboptimal results potentially due to additional learning targets they introduce. The Silhouette Coefficient of \methodname averaged over 5 runs is 0.5213, which is 77.93\% higher than the average of HDBSCAN and K-means. These quantitative results further demonstrate our method's superior capability in discerning and grouping clinical concepts and patient visits within the EHR datasets.

\vspace{0.2cm}
\subsection{Case Studies}
\label{sec:case_study}

To demonstrate the aligned clusters of clinical concepts and patient visits generated by \methodname, as well as their practical values in clinical settings, we illustrate actual clinical concepts and patient visits from our clusters of nodes and hyperedges. Specifically, we select 3 clusters from each of the two domains corresponding to the cyan, pink, and purple colors in Figure~\ref{fig:clus} (a) and (c). From the node clusters, we select the top 15 clinical concept candidates based on the clustering assignment probability \(\bm{Q}_\mathcal{V}\) from Eq.~(\ref{eq:soft_assignment}). Those are presented in the left panel of Table~\ref{tab:case}. Notably, we mark the subtype-indicative clinical concepts with color after consulting a model-blinded clinical expert. 
Similarly, from the hyperedge clusters, we select 3 patient visits based on the clustering assignment probability \(\bm{Q}_\mathcal{E}\) from Eq.~(\ref{eq:soft_assignment}), and illustrate each of them with the top 5 clinical concepts determined by the highest attention weights within the hypergraph transformer. They are shown in the right panel of Table~\ref{tab:case}. We also color the subtype-indicative clinical concepts as suggested by the clinical expert. As shown in Table~\ref{tab:case}, we can observe a notable overlap in clinical concepts between the clinical concept and patient visit clusters across all three subtypes, which validates the effectiveness of our cluster alignment in producing consistent clusters.

\begin{table*}[h]
\centering
\caption{Case studies of disease subtypes. \textnormal{
The same colors are used to indicate the correspondence with clusters in Figure~\ref{fig:clus} (\(\bm{K=5}\)). The colored clinical concepts represent the subtype-indicative ones as suggested by a clinical expert. 
Best viewed in color. }}
\includegraphics[width=\linewidth]{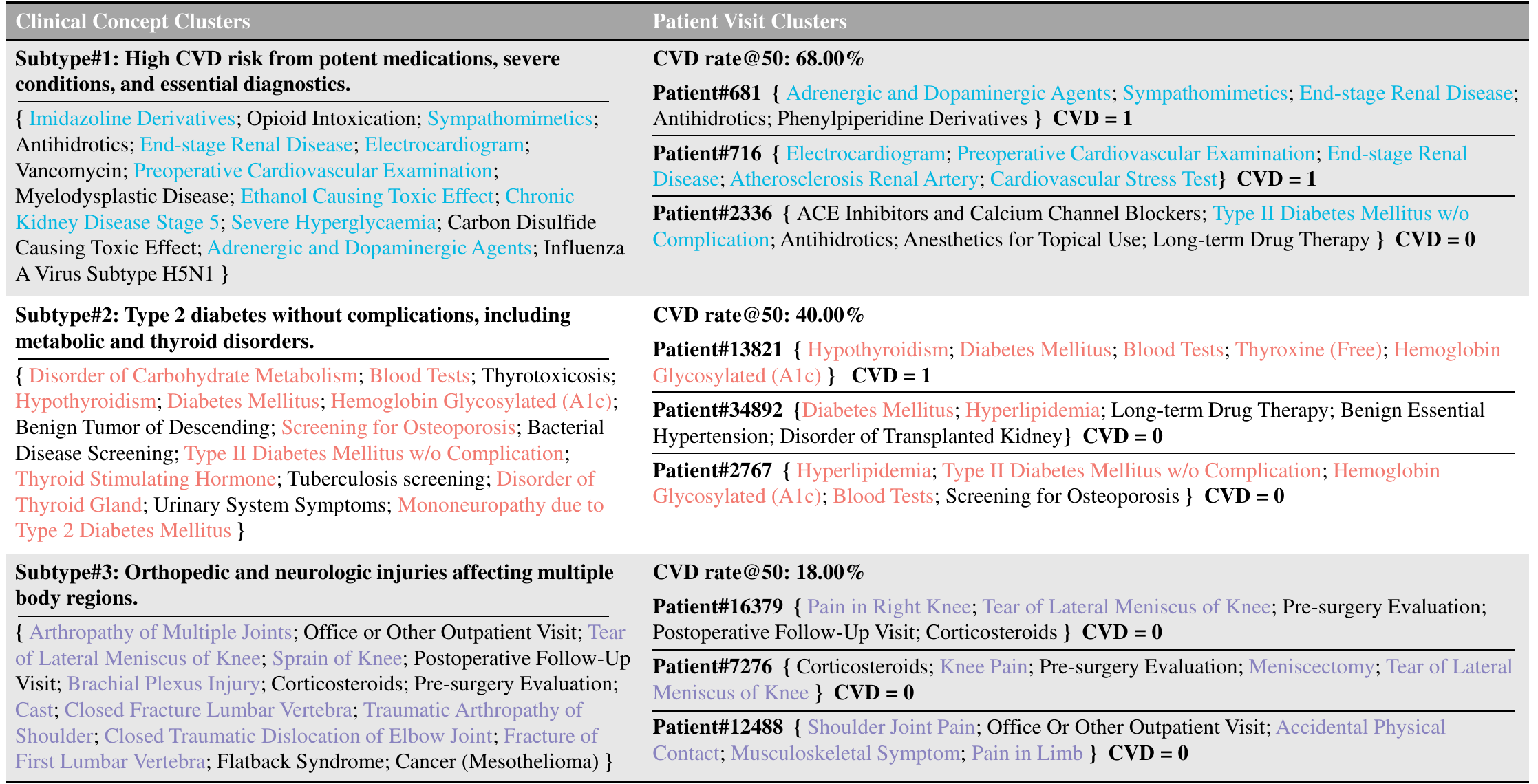}

\label{tab:case}
\end{table*}

For \textcolor{mycyan}{\textbf{Subtype\#1}}, after consulting with the clinical expert, we interpret this subtype as indicative of a heightened risk for CVD. In general, the presence of medications like \textit{Imidazoline Derivatives} and conditions such as \textit{Severe Hyperglycaemia} and \textit{End-stage Renal Disease} signals a significant cardiovascular risk \cite{head2006imidazoline, laakso1999hyperglycemia, foley1998clinical}. Diagnostic practices, including \textit{Preoperative Cardiovascular Evaluations} and \textit{Electrocardiograms}, further enhance the probability of cardiovascular disease presence in this subtype \cite{bhatia2018screening}. We identify that 34 out of the top 50 patients were confirmed to have CVD a year after their visits, which is consistent with the distribution in Figure~\ref{fig:clus} where dots of Subtype\#1 in Figure~\ref{fig:clus} (c) largely align with dots with \textcolor{red}{\textbf{CVD labels}} in Figure~\ref{fig:clus} (e). The records of Patient\#681 and \#716, who are diagnosed with CVD, also align closely with the representative clinical concepts within Subtype\#1.

\textcolor{mypink}{\textbf{Subtype\#2}} emphasizes metabolic and thyroid disorders. It delineates conditions closely associated with \textit{Type 2 Diabetes Mellitus}. Within the top 50 patients in the cluster aligning with Subtype\#2, 20 individuals are diagnosed with CVD, which indicates a comparatively moderate association with cardiovascular risk. Unlike Subtype\#1, Subtype\#2 is not marked by the use of potent pharmacological interventions and the presence of severe disease states. This demonstrates that \methodname is capable of capturing nuanced disease subtypes, thereby facilitating more targeted monitoring and intervention strategies in clinical practice for these subgroups. This advantage could potentially mitigate patients' risk of progressing to cardiovascular diseases.

\textcolor{mypurple}{\textbf{Subtype\#3}} focuses on orthopedic and neurologic injuries affecting multiple body regions, such as \textit{Sprain of Knee} and \textit{Closed Fracture Lumbar Vertebra}, with no direct ties to CVD, as evidenced by the ground truth distribution in Figure~\ref{fig:clus}(e). We also cannot observe obvious patients with CVD-related patterns. Instead, most of them share similar clinical records in musculoskeletal disorders, which are likely negatively correlated with CVD risk. This highlights that our deep co-clustering benefits from the guidance of specific disease predictions, and thereby efficiently identifies subtypes that are positively, weakly, or negatively correlated with a particular disease, such as CVD. These insights are advantageous for medical professionals in conducting precise clinical stratification and management.

\vspace{0.2cm}
\section{CONCLUSION AND DISCUSSION}

In this work, we introduce \methodname, a novel framework that jointly clusters clinical concepts and patient visits in EHR data. Specifically, we encode semantic information within the clinical concepts into a hypergraph transformer. We design a deep self-supervised co-clustering module that jointly learns a soft clustering assignment for both nodes and hyperedges. The learned clusters are then aligned through a contrastive learning objective for capturing the consistent patterns between clinical concepts and patient visits within the EHR data. Our comprehensive experiments demonstrate the superior performance of \methodname, which is 5.26\% higher than the vanilla hypergraph backbone model and 31.25\% higher than other ML baselines. Notably, \methodname is capable of discerning insightful disease subtypes related to specific diseases at different levels, enabling more targeted clinical interventions.

Currently, \methodname is in a stage of secondary data analysis. It has been tested on data from both academic benchmark MIMIC-III, and Project CRADLE, which is an actual application within the Emory Hospital Systems that provides substantial support to medical staff and researchers in the greater Atlanta and Georgia areas. Specifically, this work contributes to Project CRADLE's ongoing efforts in cardiovascular and diabetes disease management over 48 thousand patients. In a significant expansion, the method is also being adapted for use in the National Institutes of Health's All of Us\footnote{https://allofus.nih.gov/} research program. This deployment aims to harness the diverse medical records of over 38 thousand participants across the United States. Looking forward, we aim to broaden \methodname's applicational scope by extending the framework to accommodate large-scale, heterogeneous datasets that contain more modalities.

\begin{acks}
Research reported in this publication was mainly supported by the National Institute of Diabetes and Digestive and Kidney Diseases of the National Institutes of Health under Award Number K25DK135913. The research also receives partial support from the National Science Foundation under Award Number IIS-2145411, IIS-2312502, and IIS-2319449. Opinions expressed herein are those of the authors and do not necessarily represent the views of the U.S. Government. The research has also benefited from the Microsoft Accelerating Foundation Models Research (AFMR) grant program. 

\end{acks}

\newpage
\bibliographystyle{ACM-Reference-Format}
\balance
\bibliography{mybib}

\end{document}